\pdfoutput=1

\documentclass[twocolumn,english]{article}

\usepackage{times}
\usepackage[T1]{fontenc}
\usepackage[latin9]{inputenc}
\usepackage[letterpaper]{geometry}
\usepackage{float}
\usepackage{graphicx}

\makeatletter

\providecommand{\tabularnewline}{\\}

\newenvironment{lyxlist}[1]
{\begin{list}{}
{\settowidth{\labelwidth}{#1}
 \setlength{\leftmargin}{\labelwidth}
 \addtolength{\leftmargin}{\labelsep}
 }}
{\end{list}}

\makeatother

\usepackage{babel}

\begin{document}

\title{Towards a Heuristic Categorization of Prepositional Phrases in English with WordNet}

\author{Serguei Mokhov (mokhov@cs.concordia.ca)\\
Frank Rudzicz (f\_rudzic@cs.concordia.ca)\\
\$Revision: 1.14 \$}

\date{7th December 2003}

\maketitle

\begin{abstract}
\textit{This document discusses an approach and its rudimentary realization
towards automatic classification of PPs; the topic, that has not received
as much attention in NLP as NPs and VPs. The approach is a rule-based
heuristics outlined in several levels of our research. There are 7
semantic categories of PPs considered in this document that we are
able to classify from an annotated corpus.}
\end{abstract}

\section{Introduction}

Historically, prepositions have not enjoyed the attention of nouns
and verbs, being until recently relegated to the status of {}``an
annoying little surface peculiarity'' \cite{jackendoff}. However,
linguistics tells us that different syntactic categories contain distinct
semantic characteristics which are often exclusive to members of that
category \cite{levin91}. This raises two distinct yet equally important
questions: How are such categories found syntactically, and how are
their characteristics expressed semantically?

The famed linguist Sir Randolph Quirk states that {}``a preposition
expresses a relation between two entities, one being that represented
by the prepositional complement'' \cite{quirk85}.

In this paper we describe the development of a heuristics-based system.

\section{Theoretical Foundations}

Through a survey of the literature, it is clear that the study of
prepositions is becoming more intricate, and is segmented into a few
focussed areas. Our work is therefore motivated towards the construction
of a sequential system of increasingly complex levels, each of which
is representative of one of these focussed areas. The system is organized
in such a way that the product of one level becomes a dependency of
the next as outlined by the following sequence:
\begin{lyxlist}{00.00.0000}
\item [{Level~0:}] From Part-of-Speech annotated text, minimal prepositional
phrases are found at the syntactic level according to a context-free
grammar (CFG), and not categorized.
\item [{Level~1:}] Minimal prepositional phrases are augmented with a
set of labels indicating classes of semantic roles, by means of rule-based
heuristics.
\item [{Level~2:}] The proper attachment of the prepositional phrase is
attempted with shallow heuristics based on results of Levels 0 and
1, in case of ambiguity.
\item [{Level~3:}] Semantic characteristics of the PP and its co-predicate
phrases are analyzed in order to perform attachment 'intelligently',
and do discover more thorough semantic relations.
\end{lyxlist}
Each level is described in more detail in Sections \ref{sub:Discovery-of-Minimal}
through  \ref{sub:Consideration-of-Semantics}.

\subsection{\label{sub:Discovery-of-Minimal}Discovery of Minimal PPs - Syntactic
Analysis}

The first stage is to locate prepositional phrases within a text grammatically,
irregardless of semantic function. For this we turn to linguistics,
which studies PPs by their production and by their expansion as outlined
below.

\subsubsection{Production of PPs}

Prepositional phrases generally attach to noun phrases and verb phrases,
usually acting adjectively for the former and adverbially for the
latter \cite{dickphrases}, as in the example 
\begin{verse}
{\footnotesize (S}\\
{\footnotesize ~~~(NP }\\
{\footnotesize ~~~~~~(DET the )~~(HEAD boy) }\\
{\footnotesize ~~~~~~(}\textbf{\footnotesize PP}{\footnotesize{}
}\\
{\footnotesize ~~~~~~~~~in }\\
{\footnotesize ~~~~~~~~~(NP}\\
{\footnotesize ~~~~~~~~~~~~(DET the )~~(HEAD shop)
}\\
{\footnotesize ~~~~~~~~~)))}\\
{\footnotesize ~~~(VP}\\
{\footnotesize ~~~~~~is waiting}\\
{\footnotesize ~~~~~~(}\textbf{\footnotesize PP}{\footnotesize }\\
{\footnotesize ~~~~~~~~~at}\\
{\footnotesize ~~~~~~~~~(NP}\\
{\footnotesize ~~~~~~~~~~~~(DET the)~~(HEAD corner)}\\
{\footnotesize ~~~~~~~~~)))}\\
{\footnotesize )}{\footnotesize \par}
\end{verse}
At first glance, we should be able to augment grammatical noun and
verb phrases by simply adding rules of the form $NP\Leftarrow NP\, PP$
and $VP\Leftarrow VP\, PP$ \cite{jurafsky}. However this can lead
to overgeneration, especially in cases where there exist ambiguity
with regards to the part-of-speech of a potential preposition. For
example, in {}``\texttt{\footnotesize Turn off/RB the light}'',
{}``\texttt{\footnotesize off}'' is used as an adverb, but if we
are not careful with our expansion rules for $VP$, we could easily
generate {}``\texttt{\footnotesize (VP turn (PP off/IN (NP the light)))}''
if the tagging is done incorrectly. Some such errors in tagging cannot
be avoided within the scope of this project, but careful augmentation
of existing rules, such as in $NP\Leftarrow DET\, MOD\, HEAD\, PP$,
can result in a more fine-tuned grammar, as will be shown in our results
(see Section  and $VP$ expansions producing $PPs$are given in Appendix
B.

\subsubsection{Expansion of PPs}

Though there is some debate as to the semantic function played by
the prepositional phrase \cite{francez2003}, the syntactic nature
of a prepositional phraseis relatively universally accepted. It is
very safe to define the prepositional phrase specifically as havinga
preposition as its head, followed by a noun phrase or an entity whichis
always the direct object \cite{dickphrases}.

A full listing of rules for $PP$ expansions are given in Appendix
A and B.

\subsubsection{Non-minimal PPs}

We will often encounter sequences of contiguous prepositional phrases,
as in {}``\texttt{\footnotesize I saw the fool}{\footnotesize{} }\texttt{\textbf{\footnotesize on/IN}}{\footnotesize{}
}\texttt{\footnotesize the hill}{\footnotesize{} }\texttt{\textbf{\footnotesize with/IN}}{\footnotesize{}
}\texttt{\footnotesize the telescope}''. In such circumstances, we
are principally concerned with PP-attachment, which is a significant
grammatical challenge, but which also involves semantic analysis,
hence its discussion is delayed until Section \ref{sub:PP-Attachment}.
Suffice it to say, we will with some regularity encounter such a sequence
of PPs where each PP modifies the same predicative, and hence we add
the production rule $PP\Leftarrow PP\, PP$ to deal with such a circumstance.

\subsubsection{The Mechanism of Discovery}

The mechanism for discovering prepositional phrases is an implementation
of the Earley parser in Scheme. While this provides a simple interface
to grammatical tree-expansions, based purely on syntactic context-free
rules -- the algorithm is devoid of any stochastic or world knowledge,
and hence cannot be used for more complex grammatical or semantic
analysis. The instantiated grammar is an instance of a partial parser
PP-chunker which searches exclusively for prepositional phrases.

\subsection{Semantic Role Annotations \& Categorization}

Prepositions convey significant semantic relations in text, and provide
{}``the principal means of conveying semantic roles for the supporting
entities referred to in a predication'' \cite{techreport}. They
are, however, highly ambiguous - with closely related word-senses
and a relatively high degree of internal polysemy. Furthermore, definitions
as to \emph{how} semantic roles are conveyed, and \emph{how} supporting
entities relate are varied and often imprecise. At present, it is
not feasible for automatic systems to attain a degree of semantic
granularity comparable to that of collegiate dictionaries, but we
can at least categorize different uses of the prepositional phrase
by broad semantic characteristics.

\subsubsection{\label{sub:Semantic-Role-Annotations}Semantic Role Annotations}

The Penn Treebank described the prepositonal phrase as having semantic-role
subcategorizations defined by case-style relations, the 7 most frequent
of which are shown with their associated frequencies of occurrence
in Table \ref{cap:Subcategorization-of-augmented}. Although by no
means a thorough description of the semantic relations within a text,
such subcategorizations allow for a suitable indication of the manner
in which a prepositional phrase is used, and therefore to how they
modify the semantics of the phrase to which they attach.

\begin{table}[H]
\begin{centering}
\begin{tabular}{|c|c|c|}
\hline 
tag & Freq. & Description\tabularnewline
\hline
\hline 
PP-LOC & 17220 & locative\tabularnewline
\hline 
PP-TMP & 10572 & temporal\tabularnewline
\hline 
PP-DIR & 5453 & direction\tabularnewline
\hline 
PP-MNR & 1811 & manner\tabularnewline
\hline 
PP-PRP & 1096 & purpose\tabularnewline
\hline 
PP-EXT & 280 & extent\tabularnewline
\hline 
PP-BNF & 44 & beneficiary\tabularnewline
\hline
\end{tabular}
\par\end{centering}

\caption{\label{cap:Subcategorization-of-augmented}Subcategorization of augmented
PPs, ordered by frequency of occurence in the  \cite{treebank}.}

\end{table}

These semantic relations can be attached to any verb complement, but
more frequently occur with noun phrases and their clauses. The University
of Pennsylvania provides online annotated texts from SIGLEX'99 \cite{seize}
\cite{trump} with parsed examples showing each of these categories:
\begin{lyxlist}{00.00.0000}
\item [{PP-LOC:}] {}``...a federal grand jury \textbf{(PP-LOC in (NP Newark))}...''
\cite{seize}
\item [{PP-TMP:}] {}``The people who suffer \textbf{(PP-TMP in (NP the
short term))}...'' \cite{seize}
\item [{PP-DIR:}] {}``Citicorp had discussed lowering the offer \textbf{(PP-DIR
to (NP \$250 a share))}'' \cite{trump}
\item [{PP-MNR:}] {}``He could be left \textbf{(PP-MNR without (NP top-flight
legal representation))}...'' \cite{seize}
\item [{PP-PRP:}] {}``...prosecutors told Mr. Antar's lawyers that \textbf{(PP-PRP
because of (NP the recent Supreme Court rulings))}...'' \cite{seize}
\item [{PP-EXT:}] {}``AMR declined \textbf{(PP-EXT by (NP \$22.125))}...''
\cite{trump}
\item [{PP-BNF:}] {}``I baked a cake \textbf{(PP-BNF for (NP Doug))}''
\cite{treebank}
\end{lyxlist}
It is with this simple framework of 7 semantic annotations that we
choose to develop our system. Although our heuristics will be designed
to suit these categories, alternative classifications of prepositional
phrases can be used to help obtain insight towards this process.

\subsubsection{Alternative classifications}

Alternative classifications for prepositional phrases have been discussed
within the domain of semantic encoding in lexica  \cite{eagles96}
where labels are assigned to prepositions but describe whole phrases
according to whether they apply to PPs modifying predicative heads
(verbs and predicative nouns) or to PPs modifying nonpredicative heads
(nonpredicative nouns). This distinction is justified semantically
and syntactically%
\footnote{From a syntactic point of view, the choice of the prepositional label
modifying nonpredicative heads is much more restricted than for predicative
heads, although more predominantly in the romance languages than in
the germanic family. Generally, nonpredicative prepositional modifiers
'behave' more like adjectives, whereas predicative prepositional modifiers
'behave' more like adverbs. \cite{seize}%
}, and most labels associated with predicative-head modifiers coincides
directly with the semantic role annotations in Section \ref{sub:Semantic-Role-Annotations}.

The subtlety such a distinction makes is shown for modifiers of non-predicative
heads, all of which describe quality modifications. Consider, for
instance, the difference between a place-position (locative) modifier
in {}``the report is ON THE TABLE'' and a quality-place modifiers
in {}``oranges FROM SPAIN'' (origin) and {}``a road THROUGH EGYPT''
(path). Even subtle semantic differences in positional modifiers can
be noted depending on the predication of the head. Although such a
distinction is not replicated in our system, the motivation behind
the analysis of the semantic qualities of the head is applied theoretically
to our heuristics.

\subsubsection{Temporal PPs (TPPs)}

Of those semantic categorizations common across multiple paradigms,
temporal prepositional phrases have received special attention. Ian
Pratt et al. \cite{pratt97}, show that temporal prepositional phrases
with NP complements, and also with PP complements (validating our
non-minimal grammar decision!), and even sentential complements with
quantification-restrictions and structural ambiguities can be grouped
into a unified theory of \emph{generalized temporal quantifiers} which
serve as meanings for temporal PPs specifically, but also to NPs and
sentences. Their theory differs from the normal categorization theory
in that their semantics rely heavily on the $\lambda$-calculus and
warp functions, and hence is only used as a theoretical landmark. 

It should be noted that in response to Pratt et. al, Nissim Frances
et al. \cite{francez2003}%
\footnote{Who, ironically, contributed to the paper under criticism.%
}, argues that the unorthodox semantic operations and lack of syntactic
cues in that paper makes such an approach infeasible for practical
purposes, and that a simpler syntactic framework (\emph{indexical
prepositional phrases}), would better be employed, which is an approach
somewhat closer to our implementation.

\subsubsection{\label{sub:Categorization-Heuristics}Categorization Heuristics}

The heuristic approach to the problem of semantic categorization is
easily implemented in a transformational PERL script that accepts
simple PP partial parses and transforms the tag \textbf{PP} to \textbf{PP-XXX}
(where XXX is one of the annotations of Section \ref{sub:Semantic-Role-Annotations}).
Syntactic cues from the partial parse include the following:
\begin{enumerate}
\item The lexical entry of the head preposition
\item The head of the component noun phrase
\end{enumerate}
It is tempting to produce heuristics based exclusively on the preposition
alone (1), since intuitively the preposition should indicate the manner
in which a PP is being used. For instance, in the examples {}``\texttt{\scriptsize Put
the letter IN the mailbox}'', {}``\texttt{\scriptsize the dog is
IN the space capsule}'' and {}``\texttt{\scriptsize The farmer IN
the dell}'', the preposition {}``in'' is always used as a locative
preposition. A moment's reflection will reveal that this approach
is \emph{extremely} naive, because \emph{most} prepositions can be
used for multiple functions, as in the following examples:
\begin{itemize}
\item ex. {}``The transient sleeps WITHIN the cardboard box'' (LOC) vs.
{}``I'll be back WITHIN three weeks'' (TMP) 
\item ex. {}``I kicked the ball TOWARDS the net'' (DIR) vs. {}``My politics
tend TOWARDS the left'' (MNR)
\item ex. {}``I'll see you IN a couple of weeks'' (TMP) 
\end{itemize}
Our heuristics cannot in the vast majority of cases consider the preposition
alone because prepositions can often be used under different semantic
relations. However, some prepositions, such as {}``when'' or {}``because''
are so often associated with a single semantic function (temporal
and purpose, respectively), that we \emph{can} make minimal use of
the preposition's lexical entry, but not exclusively. We implement
a two-layer system of subsuming heuristics - the first which makes
a {}``guess'' at the class depending on lexical knowledge associating
words with semantic classes. For instance,
\begin{lyxlist}{00.00.0000}
\item [{TMP:}] when, until, in, during, after, before, while, under, over,
then, since, around, at, throughout
\end{lyxlist}
This will of course lead to instances where certain ambiguous prepositions
such as {}``for'' are always guessed to be of the most frequent
category - but this is excusable because of the second heuristic layer. 

The second heuristic layer subsumes (overrides decisions made in)
the first, and forms the most significant part of system and makes
use of semantic knowledge in WordNet with regards to the head of the
PP's component noun phrase in order to guide semantic classification.
Specifically, depending on the $NP$ expansion rule that expands the
direct object of the preposition, the relevant head of the noun phrase
is extracted. For instance, for pronouns and proper $NPs$, the whole
$NP$ is considered, otherwise, only the grammatic head of the noun
phrase (which has been annotated automatically) is considered.

Given the relevant head of the noun phrase, hypernyms for that word
is looked up in WordNet using the command `\texttt{\footnotesize wn
\$noun -hypen}'. This will often result in multiple possible hypernym
expansions due to the slightest polysemy of the noun phrase, so only
the first few senses%
\footnote{In WordNet, this translates as being the few most frequent senses,
as determined through corpus-based training.%
} up to some threshold (4 or 5) are considered. The hypernym trees
are then searched in decreasing frequency of sense for particular
keywords indicating semantic role, for instance the hypernym {}``\texttt{\footnotesize time
period}'', which can be derived from the noun {}``\texttt{\footnotesize yesterday}'',
indicates a temporal relation. This search is done in such a way that
the most common semantic categories are given priority.

This approach is similar to (and inspired by!) approaches using FrameNet
as the semantic resource, and an evaluation of this technique follows
in Section \ref{sub:Evaluation-of-Automatic}.

\subsection{\label{sub:PP-Attachment}PP Attachment}

Although prepositional phrases {}``... can appear within all the
other major phrase types'' \cite{manning2002}, the task of achieving
automatic PP-attachment syntactically has traditionally required attachment
either to a verb phrase or to a noun phrase, historically  accomplished
by means of a specific binary decision between two particular methods:
\begin{lyxlist}{00.00.0000}
\item [{Right~Association:}] A constituent tends to attach to another
immediately to its right. This approach favours attachment to the
noun \cite{kimball73}. \\
(ex. \texttt{\scriptsize {}``I (VP saw (NP the dog (PP with its
puppies ))''} )
\item [{Minimal~Attachment:}] A constituent tends to attach to existing
nonterminals using the fewest intermediate nodes. This approach favours
attachment to the verb \cite{frzier78}.\\
(ex. \texttt{\scriptsize {}``I (VP saw (NP the dog) (PP with
my binoculars))''} )
\end{lyxlist}
However, Whittemore et al. \cite{whittemore89}, showed that \emph{neither}
of these complementary tactics can account for more than 55\% of cases
in general texts - and that each are poor predictors of how people
resolve ambiguity \cite{mitchell2002}. The first solution to this
shortcoming is to involve statistics and lexical knowledge. It has
been shown in Brill and Resnik (1994) \cite{brill94}.{\footnotesize{}
}{}``\texttt{\footnotesize I saw the fool on the hill with the telescope}'',
where there exist numerous examples of ambiguity, but our concern
is with the ambiguity of PP-attachment. Does the prepositional phrase
{}``\texttt{\footnotesize on the hill}'' refer to the location of
the fool, or of the speaker? Does the prepositional phrase {}``\texttt{\footnotesize with
the telescop}e'' attach to the verb \texttt{\footnotesize {}``saw}'',
or to either of the nouns {}``\texttt{\footnotesize the fool}''
or {}``\texttt{\footnotesize the hill}''? 

Both of these suggested improvements to a syntactic approach to PP-attachment
- corpus-based statistical learning and world knowledge (with inference),
are beyond the scope of this project.

\subsection{\label{sub:Consideration-of-Semantics}Consideration of Semantics}

Determening semantics (Level 4) is beyond the scope of this work presentely
due to time constraints. However, we present a few general ideas on
how that could possibly be done. One (rather shallow) way is to build
subcategories of the 7 classes we have used. For example, locative-type
PPs may have lexical entries with semantic relations, such as ``part-to-whole'',
``betweenness'', and ``relative distance'' relations, which build
up ``path'' and ``orientation'' structures for ``movement-directional''
and ``stative-locational'' interpretations according to \cite{locativesem}.
Similarly to VPs, we'll have to look at the (semantics of) argument
structure of such PPs. In general, a hierarchy of subcategories of
the original classes of PPs will have to be implemented. That would
possibly require more than two passes of over the parses we currently
have. Tests for argument structure can be done as presented in \cite{verspoor-perspective}.

\section{Experiments \& Analysis}

In order to gauge the effectiveness of our heuristics-based approach,
the evaluation of our system on corpora, and the subsequent experimental
analysis must be performed. Such experiment involves a `training'
phase where the syntactic and semantic characteristics of the prepositional
phrases are analyzed and represented by heuristics, and a `testing'
phase where the efficacy of those heuristics are measured statistically.
The process is described in the following subsections.

\subsection{Experimental Setup}

\subsubsection{Corpora}

For the purposes of experiment, a collection of texts from the Wall
Street Journal (WSJ) corpus are chosen at random to be representative
of the available corpora, and are divided into two sets: The first
is comprised of \textbf{20} texts (\emph{\textasciitilde{}74\%}) used
for `\textbf{training}' of the system - that is, our heuristics are
modified to best suit these documents. The second set is comprised
of \textbf{7} texts (\emph{\textasciitilde{}26\%}) used for `\textbf{testing}'
the system - from which our empirical measures of performance are
derived. The breakdown is shown in Tables \ref{cap:Training-corpora}
and \ref{cap:Test-corpora}.

\begin{table}
\begin{centering}
\begin{tabular}{|c|c|c|c|}
\hline 
\texttt{\tiny 891027-0018.txt} & \texttt{\tiny 891027-0023.txt} & \texttt{\tiny 891027-0038.txt} & \texttt{\tiny 891027-0047.txt}\tabularnewline
\hline 
\texttt{\tiny 891027-0056.txt} & \texttt{\tiny 891027-0066.txt} & \texttt{\tiny 891027-0082.txt} & \texttt{\tiny 891027-0090.txt}\tabularnewline
\hline 
\texttt{\tiny 891027-0101.txt} & \texttt{\tiny 891027-0103.txt} & \texttt{\tiny 891027-0114.txt} & \texttt{\tiny 891027-0172.txt}\tabularnewline
\hline 
\texttt{\tiny 891027-0184.txt} & \texttt{\tiny 891030-0008.txt} & \texttt{\tiny 891030-0011.txt} & \texttt{\tiny 891030-0020.txt}\tabularnewline
\hline 
\texttt{\tiny 891030-0028.txt} & \texttt{\tiny 891030-0037.txt} & \texttt{\tiny 891030-0066.txt} & \texttt{\tiny 891030-0085.txt}\tabularnewline
\hline
\end{tabular}
\par\end{centering}

\caption{\label{cap:Training-corpora}Training documents from WSJ corpus}

\end{table}

\begin{table}
\begin{centering}
\begin{tabular}{|c|c|c|c|}
\hline 
\texttt{\tiny 891027-0007.txt} & \texttt{\tiny 891027-0040.txt} & \texttt{\tiny 891027-0081.txt} & \texttt{\tiny 891027-0099.txt}\tabularnewline
\hline 
\texttt{\tiny 891027-0111.txt} & \texttt{\tiny 891030-0019.txt} & \texttt{\tiny 891030-0093.txt} & \texttt{\tiny -}\tabularnewline
\hline
\end{tabular}
\par\end{centering}

\caption{\label{cap:Test-corpora}Testing documents from WSJ corpus}

\end{table}

\subsubsection{Human Annotation}

All 27 documents were hand-annotated for the occurence and grammatical
structure of prepositional phrases. This task was facilitated by two
factors:
\begin{itemize}
\item Only prepositional phrases (and their constituents) were annotated,
saving the annotators the trouble of doing full sentence parsing.
\item As a consequence of the first point, PP-attachment was not taken into
consideration.
\end{itemize}
Subsequently the parses were augmented with the semantic role annotations
from Section \ref{sub:Semantic-Role-Annotations}. In cases where
the prepositional phrase did \emph{not} fall into one of the 7 semantic
categories, the tag was left simply as \textbf{PP}, without additional
markup.

The annotation of the 20 training documents guided the production
of rules for grammatical expansion and heuristics for semantic role
markup. The annotation of the 7 test documents allowed for numeric
performance measures, and statistical evaluation.

\subsubsection{Measures of Performance}

Two principal measures of performance are \textbf{precision} and \textbf{recall},
defined thusly \cite{quirk85}:
\begin{lyxlist}{00.00.0000}
\item [{Precision:}] $P=\frac{\sharp\, of\, correct\, answers\, given\, by\, the\, system}{\sharp\, of\, answers\, given\, by\, the\, system}$
\item [{Recall:}] $R=\frac{\sharp\, of\, correct\, answers\, given\, by\, the\, system}{total\,\sharp\, of\, possible\, answers\, in\, the\, text}$
\end{lyxlist}
For each of these measures, an `answer' can be defined with regards
to the task at hand, and will be specified in the Section \ref{sub:Results-&-Analysis}.
For instance, when measuring the bare PP-chunker grammar, recall would
be defined as the total number of correctly identified prepositional
phrases among all actual prepositional phrases in a text.

An additional statistical measure is the F-Measure, which describes
the combined measure of precision and recall, and is described by
\[
F=\frac{\left(\beta^{2}+1\right)PR}{\beta^{2}P+R}\]
 where we consider $\beta=1$ (equal weight to precision and recall).
Non-statistical measures of performance include computational measures
and the qualitative evaluation of rules.

\subsection{\label{sub:Results-&-Analysis}Results \& Analysis}

\subsubsection{Evaluation of PP-grammar}

The discovery of minimum non-characterized PPs, which comprises the
first level of the system, is first analyzed independently, since
its performance will drastically affect the performance of later stages.

The first pass of evaluation leads to a recall of $R=\frac{302}{369}\approx81.8$\%
and a precision of $P=\frac{302}{409}\approx73.8$\%. The relatively
poor precision is quickly discovered to be a result of poor preprocessing
of the WSJ texts, which results in errors in tagging. Specifically,
erroneous prepositional phrases are found in headers of the WSJ texts
and at the ends of sentences at at points where punctuation causes
difficulty to Brill's Tagger, where errors are likely made during
the transformational tagging stage.

With additional preprocessing steps for those circumstances listed,
recall improved to $R=\frac{302}{369}\approx81.8$\% and precision
to $P=\frac{302}{381}\approx79.3$\%. Though a significant improvement,
additional improvements are possible. Analysing the training corpus,
we find that many potential prepositional phrases \emph{will not}
parse due to poor parsing of more complex noun phrases which would
normally form the object of the PP. Furthermore, the rule $PP\Leftarrow IN\, RB$,
which was originally included to deal with phrases of the form {}``\texttt{\scriptsize in
particular}'', was overgenerating many parses.

Modifying the rules for $NPs$ to take into consideration aggragate
NPs, and removing the rule $PP\Leftarrow IN\, RB$ results in a final
recall of $R=\frac{324}{369}\approx87.8$\% and precision to $P=\frac{324}{381}\approx85.0$\%,
which translates into an F-Measure of $F\approx86.4$\%.

\begin{figure}
\begin{centering}
\includegraphics[width=\columnwidth]{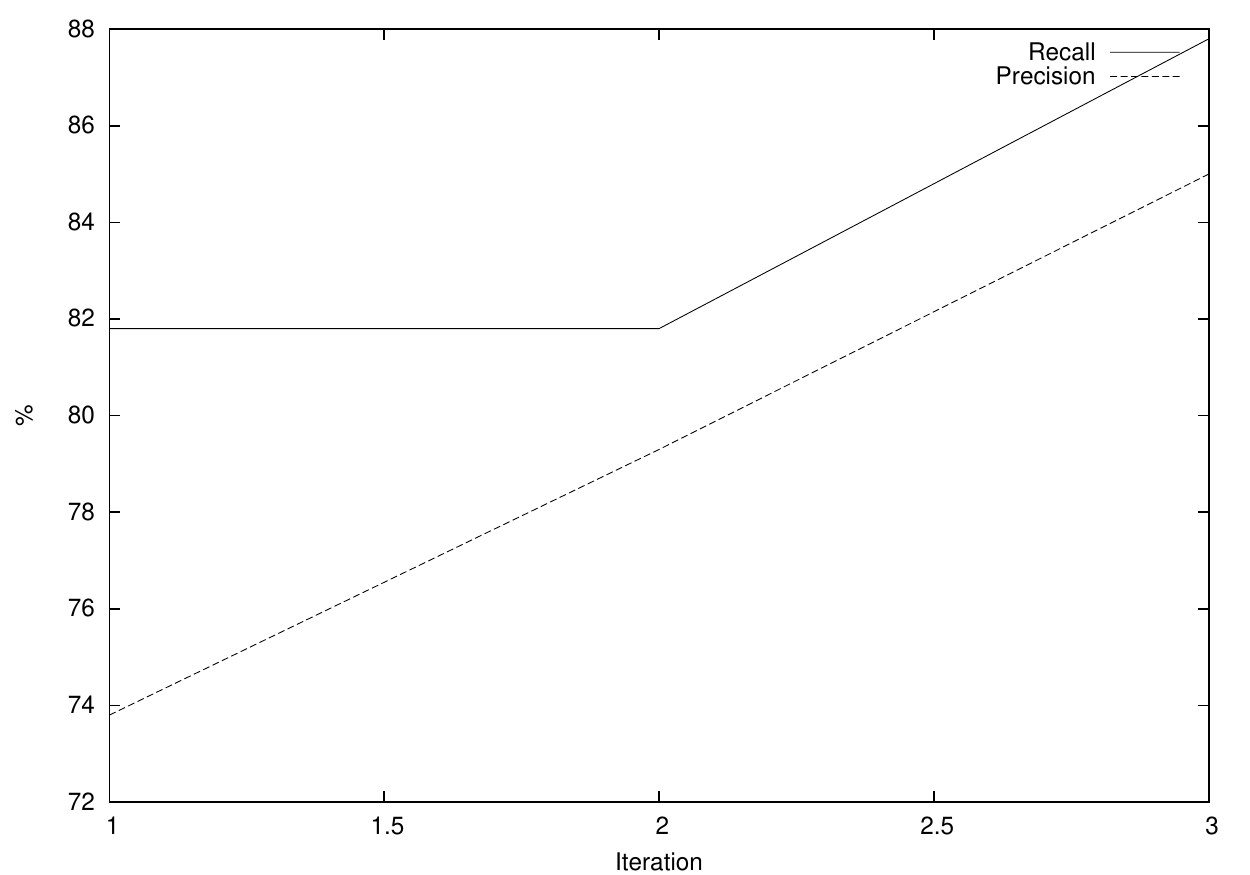}
\par\end{centering}

\caption{\label{cap:Precision-and-recall-PP-chunker}Precision and recall for
evaluation of PP-chunker.}

\end{figure}

Evidently, relatively high precision and recall can be achieved for
prepositional phrases with relatively few grammatical rules. This
is in stark contrast to the automatic noun phrase or verb phrase chunkers,which
traditionally score lower and require more numerous and more complexrules.
This tends to validate the theory in the literature which paints PPs
ashaving a basic syntactic structure, as described in Section \ref{sub:Discovery-of-Minimal}.
The improvements made at each evaluation pass are summarized in Figure
\ref{cap:Precision-and-recall-PP-chunker}.

\subsubsection{\label{sub:Evaluation-of-Automatic}Evaluation of Automatic Categorization}

Utilizing lexical knowledge first, then subsuming it with semantic
knowledge,leads to very encouraging results. Measuring the recall,
precision, and $F$-measure as previously defined for all 7 test,
we break down the analysis into the 4 most frequent catgories, where
un-annotated PPs are considered towards the total positively if they
do not strictly fit into our seven categories, and negatively otherwise%
\footnote{So total fractions do not necessarily equal the sum of its parts.%
}. This breakdown is shown in Table \ref{cap:Evaluation-of-Automatic}.

\begin{table}
\begin{centering}
\begin{tabular}{|c||c|c|c|}
\hline 
 & {\tiny Recall} & {\tiny Precision} & {\tiny $F$-measure}\tabularnewline
\hline 
{\tiny Locative PPs only} & {\tiny $\frac{128}{147}=87.1$\%} & {\tiny $\frac{128}{143}=89.5$\%} & {\tiny 88.3\%}\tabularnewline
\hline 
{\tiny Temporal PPs only} & {\tiny $\frac{78}{98}=79.6$\%} & {\tiny $\frac{78}{96}=81.3$\%} & {\tiny 80.5\%}\tabularnewline
\hline 
{\tiny Directional PPs only} & {\tiny $\frac{68}{79}=86.1$\%} & {\tiny $\frac{68}{74}=91.9$\%} & {\tiny 89.0\%}\tabularnewline
\hline 
{\tiny Manner PPs only} & {\tiny $\frac{17}{26}=65.4$\%} & {\tiny $\frac{17}{28}=60.7$\%} & {\tiny 63.1\%}\tabularnewline
\hline 
{\tiny Total} & {\tiny $\frac{293}{369}=79.4$\%} & {\tiny $\frac{293}{351}=83.5$\%} & {\tiny 81.5\%}\tabularnewline
\hline
\end{tabular}
\par\end{centering}

\caption{\label{cap:Evaluation-of-Automatic}Evaluation of Automatic Categorization}

\end{table}

It is immediately recognized that our heuristics seem exceptionally
well-suited for directional-PPs, and exceptionally poorly suited to
manner-PPs. Such a discrepancy can of course \emph{not} be linked
merely to the heuristics for these two categories, since the system
is not independent and heuristics for other categories play a significant
role with each other.

It should be noted that as more potential categories are taken under
consideration, the more error we can expect due to the relative higher
rate of perplexity.

\section{Future Work}

Most of the future work would be dedicated to the Levels 3 and 4 outlined
at the beginning, i.e. PP-attachment and Semantic Analysis of PPs.
Both problems are quite difficult to solve and will require a lot
of research-trial-and-error attempts of the existing and new proposals.
New, more robust, tools will be required, such as stochastic methods
in NLP and a lot more training and testing data. For example, we could
assign probabilities to our grammar rules and encode the attachment
information into the syntactic productions. Likewise, the detection
of semantic roles of PPs and their arguments can be done heuristically
and statistically with two or more passes.

\label{sect:bib}\bibliographystyle{alpha}
\bibliography{project}

\begin{thebibliography}{EAG96}

\bibitem[Bie95]{treebank}
A.~et.~al. Bies.
\newblock {\em Bracketing Guidelines for Treebank II Style Penn Treebank
  Project}.
\newblock Linguistic Data Consortium, 1995.
\newblock URL:
  \verb+http://www.ldc.upenn.edu/Catalog+\\\verb+/docs/treebank3/PRSGUID1.PS+.

\bibitem[BR94]{brill94}
E.~Brill and P.~Resnik.
\newblock {\em A Rule-Based Approach to Prepositional Phrase Attachment
  Disambiguation}.
\newblock COLING-94, Kyoto, Japan, 1994.

\bibitem[BR02]{mitchell2002}
Mitchell B and Gaizauskas R.
\newblock {\em A Comparison of Machine Learning Algorithms for Prepositional
  Phrase Attachment}.
\newblock University of Sheffield, 2002.

\bibitem[EAG96]{eagles96}
EAGLES96.
\newblock {\em Preliminary Recommendations on Semantic Encoding Interim
  Report}.
\newblock 1996.
\newblock URL: \verb+http://www.ilc.cnr.it/EAGLES96+\\\verb+/rep/node13.html+,
  Nov 23, 2003.

\bibitem[Frz78]{frzier78}
L.~Frzier.
\newblock {\em On Comprehending Sentences: Syntactic Parsing Strategies}.
\newblock University of Connecticut, 1978.
\newblock PhD thesis.

\bibitem[FS03]{francez2003}
N.~Francez and M.~Steedman.
\newblock {\em Categorical Grammar and the Semantics of Indexical Prepositional
  Phrases}.
\newblock Human Communication Research Centre, University of Edinburgh, 2003.

\bibitem[Hud03]{dickphrases}
R.A. Hudson.
\newblock {\em Phrases}.
\newblock November 2003.
\newblock
  \verb+www.phon.ucl.ac.uk/home/dick+\\\verb+/tta/phrases/phrases.htm#pp+.

\bibitem[Jac73]{jackendoff}
R.~Jackendoff.
\newblock {\em The Base Rules For Prepositional Phrases}.
\newblock Anderson, S.R. \& Kiparky, P. (Eds.). Holt Rinehart and Winston Inc,,
  1973.
\newblock 346-65.

\bibitem[JM00]{jurafsky}
Daniel~S. Jurafsky and James~H. Martin.
\newblock {\em Speech and Language Processing}.
\newblock Prentice-Hall, Inc., Pearson Higher Education, Upper Saddle River,
  New Jersey 07458, 2000.
\newblock ISBN 0-13-095069-6.

\bibitem[Kim73]{kimball73}
J.~Kimball.
\newblock {\em Seven Principals of Surface Structure Parsing in Natural
  Language}.
\newblock MIT Press, 1973.
\newblock Cognition 2:15-47.

\bibitem[LP91]{levin91}
B.~Levin and S.~Pinker.
\newblock {\em Lexical and Conceptual Semantics}.
\newblock Elsevier Science Publications, Amsterdam, 1991.

\bibitem[MS02]{manning2002}
C.D. Manning and H.~Schutze.
\newblock {\em Foundations of Statistical Natural Language Processing}.
\newblock MIT Press, 2002.

\bibitem[Nam95]{locativesem}
Seungho Nam.
\newblock {\em The Semantics of Locative Prepositional Phrases in English}.
\newblock University of California, Los Angeles, 1995.
\newblock URL: \verb+http://cl.snu.ac.kr/nam+\\\verb+/papers/abstract.PDF+.

\bibitem[OW02]{techreport}
T.~O'Hara and J.~Wiebe.
\newblock {\em Classifying Preposition Semantic Roles using Class-based Lexical
  Associations}.
\newblock 2002.
\newblock NMSU-CS-2002-001\\URL:
  \verb+www.cs.nmsu.edu/TechReports+\\\verb+/2002/013.pdf+.

\bibitem[PF97]{pratt97}
I.~Pratt and N.~Francez.
\newblock {\em On the Semantics of Temporal Prepositions and Preposition
  Phrases}.
\newblock University of Manchester, 1997.

\bibitem[Qui85]{quirk85}
R.~et.~al. Quirk.
\newblock {\em A Comprehensive Grammar of the English Language}.
\newblock London:Longman, 1985.
\newblock 673.

\bibitem[sei03]{seize}
November 2003.
\newblock URL: \verb+http://www.cis.upenn.edu+\\\verb+/~siglex99/seize.pred+.

\bibitem[tru03]{trump}
November 2003.
\newblock URL: \verb+http://www.cis.upenn.edu+\\\verb+/~siglex99/trump.pred+.

\bibitem[Ver]{verspoor-perspective}
Cornelia~M. Verspoor.
\newblock {\em A Perspective on PPs}.
\newblock citeseer.nj.nec.com/verspoor96perspective.html.

\bibitem[Whi90]{whittemore89}
G.~et~al Whittemore.
\newblock {\em Empirical Study of Predictive Powers of Simple Attachment
  Schemes for Post-Modifier Prepositional Phrases in Proceedings of the 28th
  Annual Meeting of the Association for Computation Linguistics}.
\newblock 1990.
\newblock pages 23-30.

\end{thebibliography}

\section{Appendix\label{appdx:pp-rules}}

\subsection{Appendix A: PP Rule Set}

The rule set is quite small and simple provided the rest of the grammar
has a quite a notion of what NP and VP are. This basic rule set is
used for Level 0 of our work. This gives us a list of unclassified
PPs from a corpus. Later on, a Perl script run to augment this information
with the 7 categories outlined in 2.2.1 keeping PP as a general classification
if we are unsure what kind of PP we are looking at. This second pass
of annotation provides classification information of Level 1 of out
work. The entire Level 0 grammar is in \texttt{pp-chunker.scm} presented
in Appendix C.The Level 1 heuristic rules are presented in Appendix
C, the \texttt{augment-pp.pl} Perl Scrip.

The below are the sample rules used:

\texttt{\scriptsize (PP (IN NP) ~~~; \textquotedbl{}on new rules\textquotedbl{},
\textquotedbl{}for him\textquotedbl{}, \textquotedbl{}by Doug\textquotedbl{} }{\scriptsize \par}

\texttt{\scriptsize ~~~~(IN IN NP) ; \textquotedbl{}because of
the rain\textquotedbl{} }{\scriptsize \par}

\texttt{\scriptsize ~~~~(PP PP) ~~~; \textquotedbl{}on new
rules for covert operations\textquotedbl{} }{\scriptsize \par}

\texttt{\scriptsize ~~~~~~~~~~~~~~~; (NB. we don't
care about attachment) }{\scriptsize \par}

\texttt{\scriptsize ~~~~(TO NP) ~~~; \textquotedbl{}to committee
officials\textquotedbl{} }{\scriptsize \par}

\texttt{\scriptsize ~~~~(IN NP VP) ; \textquotedbl{}of the top
dog running the show\textquotedbl{} }{\scriptsize \par}

\texttt{\scriptsize ~~~~~~~~~~~~~~~; (NB. spurious,
but not harmful) }{\scriptsize \par}

\texttt{\scriptsize ) }{\scriptsize \par}

\subsection{Appendix B: PP Chunker (pp-chunker.scp)}

\texttt{\scriptsize See {[}pp-chunker.scp{]}}{\scriptsize \par}

\subsection{Appendix C: PP Augmenter Script (augment-pp.pl)}

\texttt{\scriptsize See {[}augment-pp.pl{]}}
\end{document}